\documentclass[conference]{IEEEtran}
\IEEEoverridecommandlockouts
\usepackage{cite}
\usepackage{amsmath,amssymb,amsfonts}
\usepackage{algorithmic}
\usepackage{graphicx}
\usepackage{textcomp}
\usepackage{xcolor}
\usepackage{enumitem}
\usepackage{booktabs}
\usepackage{multirow}

\usepackage[linesnumbered,ruled,vlined]{algorithm2e}
\def\BibTeX{{\rm B\kern-.05em{\sc i\kern-.025em b}\kern-.08em
    T\kern-.1667em\lower.7ex\hbox{E}\kern-.125emX}}
\begin{document}

\title{RoS-KD: A Robust Stochastic Knowledge Distillation Approach for Noisy Medical Imaging\\
}

\author{
\IEEEauthorblockN{Ajay Jaiswal}
\IEEEauthorblockA{UT Austin \\}
\and
\IEEEauthorblockN{Kumar Ashutosh}
\IEEEauthorblockA{UT Austin \\}
\and
\IEEEauthorblockN{Justin F. Rousseau}
\IEEEauthorblockA{UT Austin \\}
\and
\IEEEauthorblockN{Yifan Peng}
\IEEEauthorblockA{Weill Cornell Medicine \\}
\and
\IEEEauthorblockN{Zhangyang Wang}
\IEEEauthorblockA{UT Austin \\}
\and
\IEEEauthorblockN{Ying Ding}
\IEEEauthorblockA{UT Austin \\}
}

\maketitle

\begin{abstract}
AI-powered Medical Imaging has recently achieved enormous attention due to its ability to provide fast-paced healthcare diagnoses. However, it usually suffers from a lack of high-quality datasets due to high annotation cost, inter-observer variability, human annotator error, and errors in computer-generated labels. Deep learning models trained on noisy labelled datasets are sensitive to the noise type and lead to less generalization on the unseen samples. To address this challenge, we propose a Robust Stochastic Knowledge Distillation (RoS-KD) framework which mimics the notion of learning a topic from multiple sources to ensure deterrence in learning noisy information. More specifically, RoS-KD learns a \underline{\textit{smooth, well-informed, and robust student manifold}} by distilling knowledge from multiple teachers trained on \textit{overlapping subsets} of training data. Our extensive experiments on popular medical imaging classification tasks (cardiopulmonary disease and lesion classification) using real-world datasets, show the performance benefit of RoS-KD, its ability to distill knowledge from many popular large networks (ResNet-50, DenseNet-121, MobileNet-V2) in a comparatively small network, and its robustness to adversarial attacks (PGD, FSGM). More specifically, RoS-KD achieves $>2\%$ and $>4\%$ improvement on F1-score for lesion classification and cardiopulmonary disease classification tasks, respectively, when the underlying student is ResNet-18 against recent competitive knowledge distillation baseline. Additionally, on cardiopulmonary disease classification  task, RoS-KD outperforms most of the SOTA baselines by $\sim1\%$ gain in AUC score.
\end{abstract}



\section{Introduction}
Deep learning advancements in the past decade have significantly improved the development of AI-assisted medical applications, particularly medical imaging interpretation, due to their ability to impact millions of human lives. Researchers from both academia and industry have explored several medical imaging applications such as segmentation, detection, classification, and summary generation. These applications have shown impressive, and often unprecedented potential in the assistance of healthcare specialists for preliminary diagnosis. However, the success of these applications is primarily constrained by the unavailability of high-quality, accurately annotated large training datasets. In medical imaging, dataset annotations require domain expertise,  and suffer from high inter- and intra-observer variability, human annotator error, and errors in computer-generated labels. While there has been an abundance of work around developing medical imaging algorithms~\cite{wang2018tienet,Wang_2017,jaiswal2021scalp,yao2017learning,Liu_2019_ICCV,seyyed2020chexclusion,han2021using}, handling label noise has gone largely unnoticed. Many recently proposed studies have identified that label noise can significantly impact the performance of deep learning models which can have catastrophic implication in the medical domain, considering its direct association with safety of human lives \cite{algan2020label,jaiswal2021radbert}.

\begin{figure}[h]
\begin{center}
   \includegraphics[width=\linewidth]{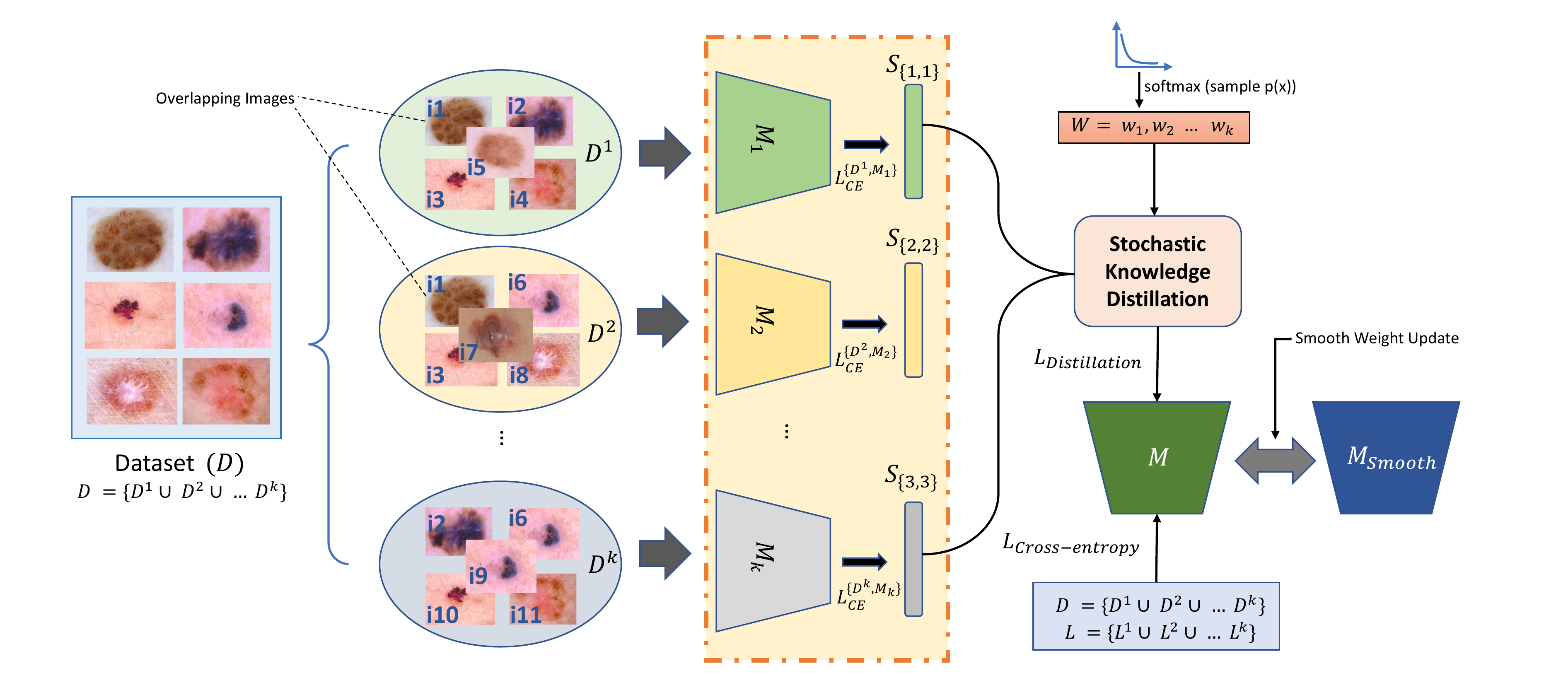}
\end{center}
   \caption{Model architecture of our Robust Stochastic Knowledge Distillation Framework. In our approach, the training dataset is divided into overlapping subsets and multiple teachers are trained using them. Knowledge from each teacher is distilled using our stochastic knowledge distillation module. To ensure the smooth update by any participating teacher, our model $M_{smooth}$ is updated by averaging multiple checkpoints along the training trajectory.}
\label{fig:1}
\end{figure}

Due to its resource-intensive nature, the challenge of labeling a large volume of medical images has  encouraged researchers to use automated tools with weak supervision. For example, many publicly available large radiology datasets  such as NIH ChestX-rays and MRI, MIMIC-CXR, and OpenI are labelled using NLP-based label extraction tools on radiology reports. To handle label noise with high variability, we borrow fundamental ideas from ensemble learning and knowledge distillation (KD), and propose a novel ensemble-based Robust Stochastic KD framework: RoS-KD. In contrast to traditional ensemble-based KD approaches  in which all participating teachers learn from the same training dataset \cite{Fukuda2017EfficientKD,Wu2019MultiteacherKD,Yang2020ModelCW}, RoS-KD unprecedentedly divides the training dataset into overlapping subsets which allow each participating teacher to spot unique noise patterns. Overlapping allows teachers to not be in complete disagreement, while unique subsets of training data help teachers break the symmetry of learning similar noise patterns. Motivated by the work of \cite{li2017visualizing}, we propose to incorporate additional smoothening in the knowledge distillation step of RoS-KD  which helps in flattening the global minima during the optimization, improving generalization.

RoS-KD is inspired by a classroom scenario, where a student learns the same concept from multiple teachers (sharing core fundamentals along with individual noisy knowledge), to be able to segregate noisy knowledge from the core fundamentals. Our extensive experiments on two popular medical classification tasks illustrate the superior performance of RoS-KD with respect to several recent competitive KD baselines. Additionally, we provide ablation to understand the importance of learning from overlapping datasets.  To ensure that the student should learn by assimilating the knowledge from multiple teachers instead of abruptly updating itself by a single teacher, we propose a smooth parameter update using weight averaging. Furthermore, we show that RoS-KD based students are more robust to adversarial attacks. To the best of our knowledge, we find that this is the first work to study the adversarial robustness perspective of knowledge distillation in the medical imaging domain using two challenging real-world datasets of cardiopulmonary disease classification and lesion classification. 
Our main contributions can be summarized as:

\begin{itemize}
    \item We propose a novel stochastic knowledge distillation framework (RoS-KD) which distills knowledge from multiple teacher networks trained on \textbf{\textit{overlapping subsets of noisy labelled data}} and \textbf{\textit{dynamically assign weights}} to teacher models to enhance deterrence to noise and improve generalization on unseen data during inference. 
    \item We propose to use \textbf{\textit{smooth parameter averaging update}} with our novel knowledge distillation framework to effectively moderate any abrupt learning by the student.
    \item Our extensive experiments on two popular \textit{real-world dataset based} medical classification tasks show the performance benefit of RoS-KD over several baseline methods. More specifically, RoS-KD achieves $>2\%$ and $>4\%$ improvement on F1-score for lesion classification and cardiopulmonary disease classification task respectively, when underlying student is ResNet-18. Additionally, on cardiopulmonary disease classification task, RoS-KD outperforms most of the state-of-the-art baselines by $\sim1\%$ gain in AUC score.
    \item We experimentally verify that RoS-KD produces students which are \textbf{\textit{highly robust}} to adversarial attacks compared to different baseline methods. RoS-KD students also have \textbf{\textit{highly smooth loss landscape}}, which can explain their better generalization capability on unseen data.
\end{itemize}

\section{Background Work}
Knowledge Distillation (KD) \cite{hinton2015distilling} is an effective way to compress large models into smaller ones with comparable performance. KD is based on a teacher-student learning paradigm in which the student learns from the soft-targets of the teacher network. Recently, some methods employ multiple teachers and show great promises in further boosting student model performance effectively \cite{Fukuda2017EfficientKD,Wu2019MultiteacherKD,Yang2020ModelCW}. Most of these existing methods using multiple teachers simply assign  equal weight to all teacher models during
the whole distillation process. Moreover, they primarily use the same training data to train each participating teacher network which make all teachers prone to learning the similar noise pattern available in the training data. Recently, \cite{yuan2021reinforced} identified that  individual teacher models may perform differently on different data-points due to optimization strategy and parameter initialization. This encourages us to assign different weights to teacher models for different training instances during training. RoS-KD  provides a noise-tolerant perspective of knowledge distillation and introduces smoothening as a key to improving its performance.  Compared to previous knowledge distillation methods using multi-teacher models usually fix the same weight for a teacher model on all training examples, RoS-KD design allows stochastic assignment of weights to each participating teacher models for each training example during training.


\section{The Proposed Approach}
\subsection{Model Architecture}
A schematic representation of our RoS-KD framework is given in Figure \ref{fig:1}. The core RoS-KD framework is based on ensemble-based knowledge distillation where multiple teacher networks are used to teach the student network. In RoS-KD, the training data $D$ is divided into $k$ overlapping subsets with an overlap ratio of $p\%$ such that $D = \{D^1 \cup D^2 ... \cup D^k\}$. We train $k$ different model architectures $\{ M^1_{D^1}, M^2_{D^2}, ... ,M^k_{D^k}\}$ corresponding to each $D^i$ using cross-entropy loss. We use  the overlapping dataset instead of a traditional non-overlapping complete dataset $D$ to ensure that each teacher has sufficient agreement on the common representation of $D$ along with its unique share of disagreement due to noise. This setting mimics the situation that each teacher knows the same topic in its own unique style. Our student network $M$ is trained using our stochastic knowledge distillation module which combines the soft-labels generated for each training example $x_i \in D$ by an individual teacher. To mitigate the abrupt impact of any teacher in the student's learning process, we propose to average multiple checkpoints of $M$ along the training trajectory to update $M_{smooth}$, which is our final student network (RoS-KD). Smooth parameter averaging is extremely easy to implement, which can improve standard generalization of students, and has almost no computation overhead.

\subsection{Robust Stochastic Noise-Tolerant Knowledge Distillation}
Many large-scale medical imaging datasets \cite{johnson2019mimic,Wang_2017} are labelled using automated tools under the weak supervision of domain experts and have highly variable noise across data samples. Many recent studies have shown that label noise can significantly impact the performance of deep learning models and lead to degraded generalization. Our robust stochastic KD (RoS-KD) framework is motivated by the idea that teachers know not only the common fundamental details of the topic but also some unique explanations. Therefore, if a student learns from multiple teachers, it enables the student to learn multiple unique explanations of the topic along with the common fundamentals of the concept. This will help the student be better than individual teachers due to the diversity of information the student has learned along with identifying any conflicting information about the same topic. RoS-KD incorporates this setting by proposing to divide the training dataset into overlapping blocks and training multiple teachers architectures (e.g., ResNet-18, DenseNet-121, and MovileNet-V2) on the overlapping datasets. RoS-KD differs from the conventional ensemble KD approaches which use the same dataset to train each participating teacher network, because RoS-KD allows the teacher network to be consistent with each other and learn additional unique information.

Next, we aggregate knowledge from the multiple trained teachers using stochastic weighted distillation. In each iteration, we randomly sample the weight of each teacher from an exponential distribution. The weight is used to decide the teacher's contribution for updating student $M$ in that iteration. This aggregation process simulates that the student learns the knowledge from one teacher and compares it to others.  Our stochastic weighted distillation ensures that only one teacher plays a significant role in the update at one time.  
Therefore, we do not need to jointly minimize KL divergence among multiple teachers with an equal contribution at the same time. Furthermore, to improve the generalization capability of our RoS-KD framework, we propose to use smooth parameter averaging update  to effectively moderate any abrupt learning by the student. This ensures that no teacher will be allowed to make significantly large updates which helps students to relax, think, and update gradually.

\subsubsection{Smooth Parameter Averaging Update} 

In our RoS-KD framework, we use a smooth parameter averaging update to improve the generalizability. The update can effectively moderate any abrupt learning by the student, thus ensures that no teacher will make a significantly large update.

It is widely believed that the loss surface at the final learned weights for well-generalized models is relatively ``flat"\cite{li2017visualizing}. To ensure the smooth update of our final student model $M_{smooth}$, we propose to enforce weight smoothness, by averaging multiple checkpoints along the training trajectory. Our parameter averaging update can be interpreted as approximating the fast geometric ensembling \cite{garipov2018loss}, by aggregating multiple checkpoint weights at different training times \cite{izmailov2018averaging}. 
\begin{align}
    \mathcal{W}^T_{M_{smooth}} &= \frac{\mathcal{W}^{T-1}_{M_{smooth}} \times n  + \mathcal{W}^{T}_{M}}{ n + 1}\\
    \mathcal{W}^{T}_{M} &= \mathcal{W}^{T-1}_{M} + \Delta \mathcal{W}^{T}_{M}
\end{align}
where T indexes the training epoch, $n$ is the number of past checkpoints to be averaged, $\mathcal{W}_{M_{smooth}}$  denotes the averaged network weight, $\mathcal{W}_{M}$ represents the current network weight, and $\Delta \mathcal{W}_{M}$ indicates the SGD update.

The Smooth Parameter Averaging Update provides an opportunity to make the student network to be robust and learn flatter solutions. Smooth parameter averaging is straightforward  to implement with almost no computational overhead.

\subsubsection{Loss Function} 
For a $C$-class classification task, given a teacher network $m_i=M^i_{D^i}$ trained on a data subset $D^i$ and input $x^i$, we leverage the logit $z^i \in R^C$ (final output before the softmax layer) from $m_i$ to supervise the desired student network $M$. Following the setting of knowledge distillation, the logit $z^i$ is distilled to the knowledge $q^i \in R^C$ by the temperature $\tau$ according to the following:
\begin{equation}
q^i_j = \sigma_{\tau} (z^i_j) = \frac{\exp(z^i_j / \tau)}{\sum_{j = 1}^C \exp (z^i_j / \tau)}    
\end{equation}
where $q^i_j$ denote the $j$th element of $q^i$ and $\sigma_{\tau}(.)$ represents the standard softmax function with the distilling temperature $\tau$. Usually, $\tau$ is a positive value greater than $1$, and a higher value for $\tau$ can produce a softer probability distribution over classes.

Next, we sample weights from the exponential distribution for $K$ participating teacher models $\{m_1,...,m_K\}$ as $\{w_i, ..., w_K\}$. Then, we supervise the student network $M$ by minimizing the following mini-batch loss $l$ over $L$ samples:

\begin{equation}
    l = \sum_{i=1}^L\{\alpha\tau^2 \sum_{m=1}^K \{w_m.\mathcal{KL}(q^i_{m_i}, p^i)\} + (1- \alpha)\mathcal{CE}(M(x^i), y^i)\}
\end{equation}
Here, $\mathcal{KL}$ and $\mathcal{CE}$ represent KL divergence and cross-entropy loss, respectively. 
$p^i = \sigma_{\tau}(M(x^i))$. $y^i$ is hard label of $x^i$. The hyper-parameter $\alpha \in [0,1]$ balances the KL divergence and the cross-entropy loss. $\tau$ is a specified temperature.

\section{Experimental Results}

\subsection{Task Formulation} While in principle, our method of multi-teacher learning should be applicable to any deep learning task, we restrict our focus to problems where the dimension of the output vector is small. Such a constraint will remove tasks that require intensive training for all the teacher networks, for example, in tasks such as segmentation, localization. Thus, we restrict our experiment and analysis to classification tasks to keep focus on understanding the benefits of our design. In this work, we evaluate the effectiveness of our RoS-KD framework on two popular medical imaging classification tasks - lesion classification and cardiopulmonary disease classification.
The skin lesion dataset consists of 25,331 skin lesion images divided into eight different clinical scenarios \cite{zhou2019multi}. We aim to build a model to classify an image into one of the eight clinical scenarios. 
The NIH Chest X-ray dataset consists of 112,120 chest X-rays collected from 30,805 patients, and each image is labeled with 8 cardiopulmonary disease labels \cite{Wang_2017}. We followed the same protocol as \cite{li2018thoracic}, to shuffle our dataset into three subsets: 70\% for training, 10\% for validation, and 20\% for testing. In order to prevent data leakage across patients, we ensure no overlap within our train, validation, and test set.


\subsection{Experimental Settings}
To prove the efficacy of our RoS-KD framework, we have selected two popular medical imaging tasks - lesion classification and cardiopulmonary disease classification. For both tasks, we divide the training split of the dataset into 5 overlapping subsets (with overlap ratio 0.4) and train 5 teacher networks (ResNet-18,34,50, MobileNet-v2, and DenseNet-121) using an SGD optimizer with a momentum of 0.9 and weight decay of $2e^{-4}$. The initial learning rate is set to 0.1, and the networks are trained for 50 epochs with
a batch size of 64. The learning rate decays by a factor of 10 at the 25th and 40th epoch during the training. For all our experiments, we have kept temperature hyperparameter $\tau = 0.5$. We set $\alpha = 0.9$ to regulate the weight between the distillation and cross-entropy loss during RoS-KD training.  Additionally, we provide an initial warmup of 10 epochs to $M_{smooth}$ during the smooth parameter averaging update. All our models are trained using 4 Quadro RTX 5000 GPUs. 

\subsection{Baselines}
\label{baselines}
\subsubsection{Baseline I} RoS-KD proposes a noise-tolerant stochastic knowledge distillation framework which distills knowledge from multiple teacher networks in a student network. For evaluation of RoS-KD, we have selected ResNet-18 architecture as the default student network. Our first baseline is a standard ResNet-18 architecture trained on cardiopulmonary and lesion classification task.
\subsubsection{Baseline II} Our second baseline follows the standard knowledge distillation setting proposed in \cite{hinton2015distilling} and train the ResNet-18 student network with the assistance of comparatively larger teacher network DenseNet-121.
\subsubsection{Baseline III} We implemented a multi-teacher ensemble model similar to \cite{Yang2020ModelCW} where every teacher model is assigned an equal weight in KD, and the student model (ResNet-18) learns from an aggregated distribution by averaging teacher outputs.
\subsubsection{Baseline IV} Recently, \cite{yuan2021reinforced} proposed reinforcement based method to perform adaptive weight assignment to each participating teachers in a multi-teacher learning framework. We adapted their method for our tasks, and \textit{surprisingly} found that RoS-KD which randomly sample the weights from the exponential distribution for each teacher, can significantly outperform their computationally inefficient RL-based design.  
\subsubsection{Baseline V} Our baseline IV is the RoS-KD framework which only uses overlapping subset of training data along with stochastic importance to individual participating teacher network. Note that this baseline \textbf{doesn't use smooth parameter averaging}  during distillation.


\subsection{Results and Discussion}
\subsubsection{Lesion Classification Task}
The lesion classification task is a one-class classification problem where RoS-KD assigns one class to each input image among 8 class categories. Table \ref{table:classification} presents the performance comparison of  RoS-KD with respect to several baseline methods explained in Section \ref{baselines}. RoS-KD achieves a significant performance gain of $+3.2\%$ in F1-score over traditional single teacher based KD framework (Baseline II). It addition, it also outperforms fixed weight multi-teacher KD framework (Baseline III) by $+2.2\%$. Note that Baseline III uses the exact same set of teacher architectures and training hyperparameters for fair comparison. Surprisingly, RoS-KD beats recently published RL-based dynamically weighted baseline \cite{yuan2021reinforced} significantly by $+3.9\%$. Moreover, when compared to the performance of a standard network (ResNet-18), RoS-KD based ResNet-18 model achieves $+5.8\%$ better F1-score. In order to investigate the performance consistency of RoS-KD across different student architectures, we experimented with popular ResNet-18/34/50, MobileNet-v2, and DenseNet-121 as students. 
Noticeably, for DenseNet-121 and ResNet-34, RoS-KD achieves $+3.2\%$ and $+3.1\%$ gain in F1-score respectively.  

\begin{table}[h]
\centering
\scriptsize
\begin{tabular}{lcccccc} 
 \toprule
  \multirow{2}{*}{\textbf{Settings}} & \multicolumn{3}{c}{\textbf{Lesion Classification}} & \multicolumn{3}{c}{\textbf{Cardiopulmonary Classification}} \\ 
 \cmidrule(rr){2-4}\cmidrule(rr){5-7}
 & Precsion & Recall & F1 & Precision & Recall & F1 \\
 \midrule
 Baseline I & 0.653 & 0.664 & 0.658 & 0.298 & 0.301 & 0.299\\
 Baseline II & 0.680 & 0.692 & 0.684 & 0.312 & 0.348 & 0.329\\
 Baseline III & 0.691 & 0.704 & 0.694 & 0.300 & 0.316 & 0.308 \\ 
 Baseline IV & 0.683 & 0.669 & 0.677 & 0.304 & 0.310 & 0.307 \\ 
 Baseline V  & 0.703 & 0.714 & 0.705 & 0.341 & 0.327 & 0.334 \\

 \midrule
 \textbf{RoS-KD} &  \textbf{0.713} & \textbf{0.726} & \textbf{0.716} & \textbf{0.360} & \textbf{0.339} & \textbf{0.349} \\
 
 \bottomrule
 \vspace{0.05em}
\end{tabular}
\caption{Performance Comparison of RoS-KD with respect to baselines on the lesion and cardiopulmonary classification task.}
\vspace{-0.6cm}
\label{table:classification}
\end{table}



\begin{table}[h]
\centering
\tiny
\begin{tabular}{lllcccccc} 
 \toprule
 \multirow{2}{*}{\textbf{Dataset}} & \multirow{2}{*}{\textbf{Attack}} & \multirow{2}{*}{\textbf{Settings}} & \multicolumn{3}{c}{\textbf{Before Attack}} & \multicolumn{3}{c}{\textbf{After Attack}} \\ 
 \cmidrule(rr){4-6}\cmidrule(rr){7-9}
 & & & Precsion & Recall & F1 & Precision & Recall & F1 \\
 \midrule
 Lesion & PGD & Baseline III & 0.691 & 0.704 & 0.694 & 0.363 & 0.293 & 0.309\\
 & & RoS-KD & 0.713 & 0.726 & 0.716 & \textbf{0.417} & \textbf{0.359} & \textbf{0.354}\\ 
 \cmidrule{2-9}
 & FSGM & Baseline III & 0.691 & 0.704 & 0.694 & 0.445 & 0.377 & 0.385\\
 & & RoS-KD & 0.713 & 0.726 & 0.716 & \textbf{0.475} & \textbf{0.422} & \textbf{0.447}\\ 
 
 \midrule
 
 Cardio. & PGD & Baseline III & 0.312 & 0.348 & 0.329 & 0.150 & 0.131 & 0.139\\
 & & RoS-KD & 0.360 & 0.339 & 0.349 & \textbf{0.189} & \textbf{0.175} & \textbf{0.182}\\ 
\cmidrule{2-9}
 & FSGM & Baseline III & 0.312 & 0.348 & 0.329 & 0.162 & 0.144 & 0.152\\
 & & RoS-KD & 0.360 & 0.339 & 0.349 & \textbf{0.201} & \textbf{0.187} & \textbf{0.194}\\ 
 \bottomrule
 \vspace{0.05em}
\end{tabular}
\caption{Performance comparison of RoS-KD with varying student architecture wrt. Baseline III. Note that we have used exactly same teacher models for KD in both Baseline III and RoS-KD for fair comparison. Norm is $l_2$. Radius $\epsilon = \frac{128}{255}$.}
\label{table:adversarial}
\vspace{-0.5cm}
\end{table}

\subsubsection{Cardiopulmonary Disease Classification Task}
The cardiopulmonary disease classification task is a multi-class classification problem. RoS-KD assigns one or more labels among 8 cardiopulmonary classes. Table \ref{table:classification} presents the performance comparison of RoS-KD with respect to several baseline methods explained in Section \ref{baselines}. RoS-KD achieves a significant performance gain of $+2.0\%$ in F1-score over traditional single teacher based KD framework (Baseline II). It addition, it also outperforms fixed weight multi-teacher KD framework (Baseline III) by $+4.1\%$. Note that Baseline III uses the exactly same set of teacher architectures and training hyperparameters for fair comparison. Moreover, when compared to the performance of a standard network (ResNet-18), RoS-KD based ResNet-18 model achieves $+4.5\%$ better F1-score. Finally, in comparison with \cite{yuan2021reinforced}, which uses dynamic weight assignment (Baseline IV), RoS-KD archives $2.7\%$ better performance. To investigate the performance consistency of RoS-KD across different student architectures, we experimented with popular ResNet-18/34/50, MobileNet-v2, and DenseNet-121 as students. It can be clearly observed that RoS-KD performs significantly better across all student architecture. Noticeably, for MobileNet-v2 and ResNet-18, RoS-KD achieves $+3.8\%$ and $+4.1\%$ gain in F1-score respectively.  

Unlike the lesion classification task, cardiopulmonary disease classification task is comparatively well-studied by the medical-imaging community and there exists many well established baselines \cite{wang2017chestx, wang2018tienet, yao2017learning, rajpurkar2017chexnet, kumar2018boosted, Liu_2019_ICCV, seyyed2020chexclusion} to evaluate the performance of newly proposed algorithms. We compare RoS-KD performance with reference models, which have published state-of-the-art performance of disease classification on the
NIH dataset \cite{wang2017chestx}. We have used Area under the Receiver Operating Characteristics (AUC) to estimate the performance of our
RoS-KD in Table \ref{table1}. Our results also present the 3-fold cross-validation to show the robustness of our reported AUC scores. Compared to
other baselines, RoS-KD achieves a mean AUROC score of 0.838 using DenseNet-121 across the 8 different classes, which is $1\%$ higher than the best performing baseline on disease classification.

\begin{table*}[h]
\centering
\begin{tabular}{lccccccccc}
\toprule
\textbf{Method} & \textbf{Atel.} & \textbf{Cardio.} & \textbf{Effus.} & \textbf{Infilt.} & \textbf{Mass} & \textbf{Nodule} & \textbf{Pneum.} & \textbf{Pneumo.} & \textbf{Mean}\\
\midrule
\textit{Wang et. al.}\cite{wang2017chestx} &  0.72 & 0.81 & 0.78 & 0.61 & 0.71 & 0.67 & 0.63 & 0.81 & 0.718\\
\textit{Wang et. al.}\cite{wang2018tienet} &  0.73  & 0.84  & 0.79    & 0.67  & 0.73  &  0.69 &  0.72  &  0.85 & 0.753\\
\textit{Yao et. al.}\cite{yao2017learning} &  0.77 & 0.90 & 0.86 & 0.70 & 0.79 & 0.72 & 0.71 & 0.84 & 0.786\\
\textit{Raj. et. al.}\cite{rajpurkar2017chexnet} & 0.82 & 0.91 & 0.88 & 0.72 & 0.86 & 0.78 & 0.76 & 0.89 & 0.828\\
\textit{Kum. et. al.}\cite{kumar2018boosted} & 0.76 & 0.91 & 0.86 & 0.69 & 0.75 & 0.67 & 0.72 & 0.86 & 0.778\\ 
\textit{Liu et. al.}\cite{Liu_2019_ICCV} & 0.79 & 0.87 & 0.88 & 0.69 & 0.81 & 0.73 & 0.75 & 0.89 & 0.801\\
\textit{Seyed et. al.}\cite{seyyed2020chexclusion} & 0.81 & 0.92 & 0.87 & 0.72 & 0.83 & 0.78 & 0.76 & 0.88 & 0.821\\ 
\midrule
\textbf{RoS-KD (Ours)}             & 0.83& 0.91& 0.89& 0.77& 0.85& 0.78& 0.79& 0.88 & 0.838\\
\hspace{2em}(std) & $\pm 0.00$ & $\pm 0.01$ & $\pm 0.01$ & $\pm 0.01$ & $\pm 0.02$ & $\pm 0.00$ & $\pm 0.01$ & $\pm 0.02$\\
\bottomrule
\vspace{0.05em}
\end{tabular}
\caption{\label{table1}
Comparison with the baseline models for AUC of each class and average AUC (three independent runs).
}
\label{tab:classificationresult}
\vspace{-0.8cm}
\end{table*}

\begin{figure}[htbp!]
\centering
\vspace{-0.3cm}
\includegraphics[width=\linewidth]{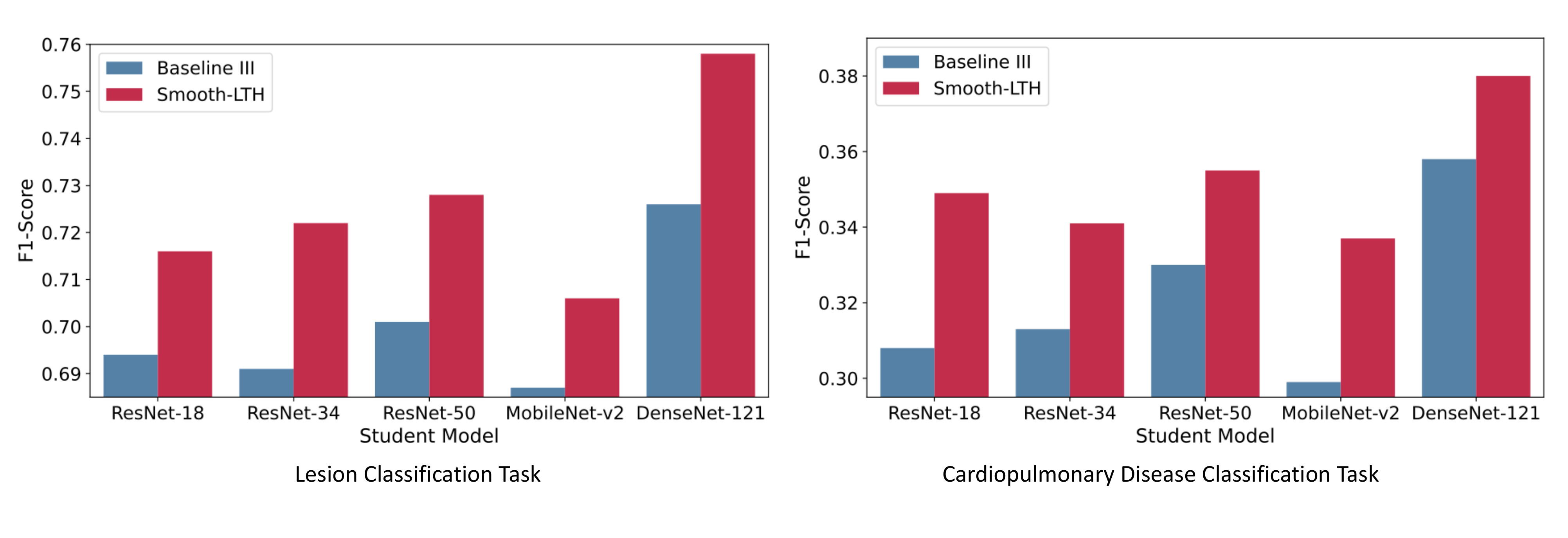}
\vspace{-0.8cm}
\caption{Performance comparison of RoS-KD with varying student networks wrt. Baseline III on lesion and cardiopulmonary disease classification task.}
\vspace{-0.3cm}
\label{fig:cardio}
\end{figure}

\subsubsection{How does overlapping impact the performance?} 
\begin{figure}[h!]
\centering
\includegraphics[width=\linewidth]{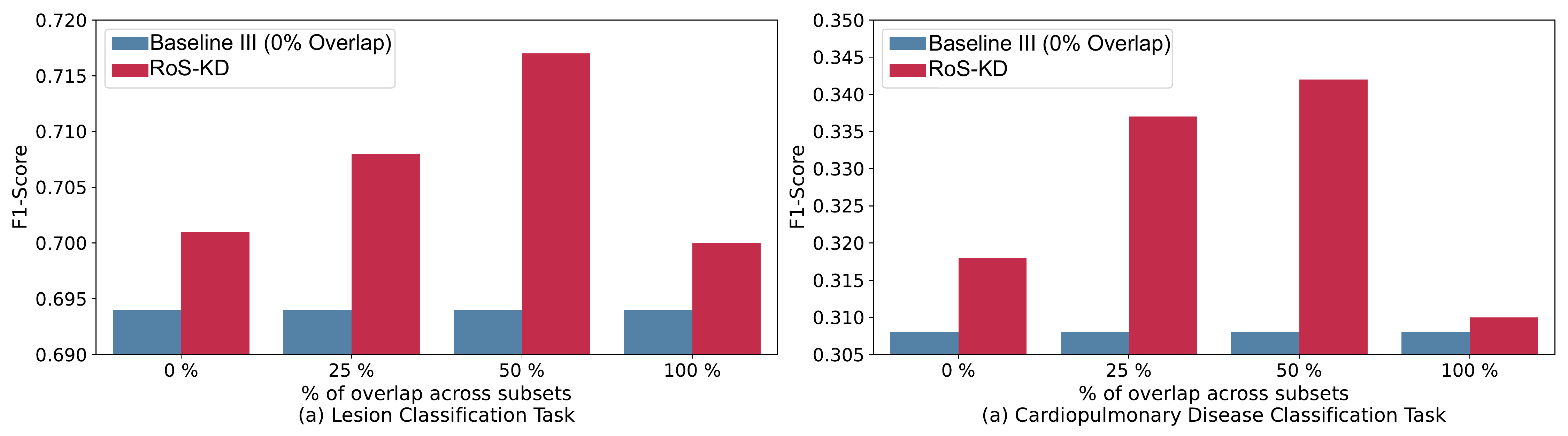}
\caption{Impact of overlapping ratio in training subsets of teacher models  trained with RoS-KD. Baseline III doesn't use any overlapping training subset.}
\vspace{-0.6cm}
\label{fig:comparison_overlap}
\end{figure}

One key contribution of this work is to identify the hidden gem to use overlapping subsets of training data to train individual teacher networks in a Multi-Teacher Knowledge Distillation Framework. Figure \ref{fig:comparison_overlap} illustrates the performance comparison of RoS-KD trained with overlapping subsets in comparison with Baseline III. An overlap ratio of $0\%$ imply that the training subsets are disjointed while overlap ratio of $100\%$ implies that all teachers are trained using exactly the same data. We observed that distillation with 0\% overlap have comparatively better performance than 100\% overlap. Based on our empirical observations, we argue that with 0\% overlap, each individual teacher will attempt to learn its own discriminative features which have unique properties compared to other teachers, and these features can add significant value to the student learning. To ensure minimal disagreement among teachers trained on disjointed subsets, we investigated how sharing training samples across the teachers will impact the RoS-KD performance. For both lesion and cardiopulmonary disease classification task, we observed that overlapping significantly improves the performance of RoS-KD. 

\begin{figure*}[h!]
\centering
\includegraphics[width=0.8\linewidth]{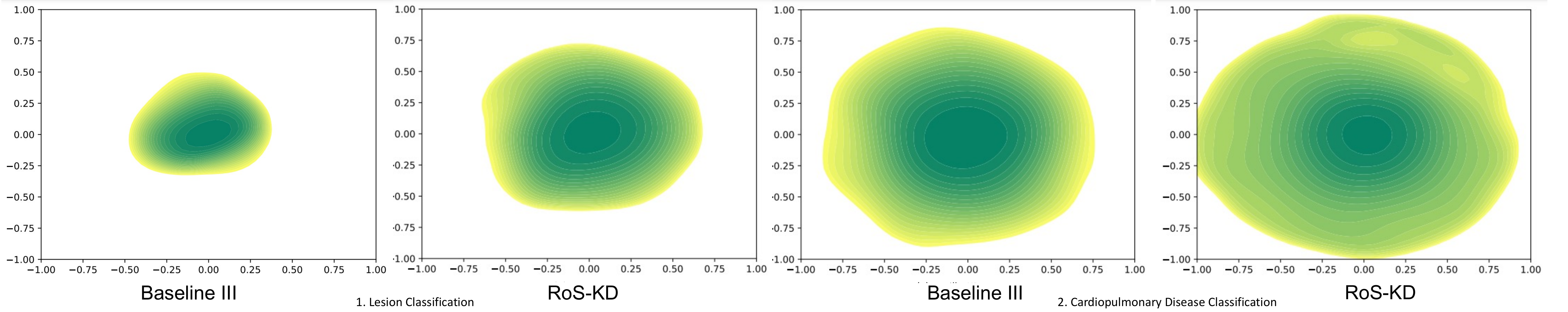}
\caption{Comparison of loss landscape of models trained with Baseline III and RoS-KD framework for lesion classification and cardiopulmonary disease classification task. Loss plots are generated with the same original images randomly chosen from the test dataset for Baseline III and RoS-KD. Z-axis denote the loss value clamped at 8.0 for better visualization. We choose Baseline III because of its best performance.}
\label{fig:loss_landscape}
\vspace{-0.5cm}
\end{figure*}

\subsection{Adversarial Robustness}
AI-assisted medical imaging can be used to make critical medical decisions and directly impact patient life. Recently, adversarial attacks have received significant attention in which an adversary tries to malice the AI-classifier by adding a small magnitude of noise to change its prediction \cite{kurakin2016adversarial}. Considering high stakes of medical imaging in clinical decision-making, it is very important to ensure that AI-algorithms are robust to adversarial attacks. Table \ref{table:adversarial} presents the robustness of RoS-KD in comparison to Baseline III under two representative attacks: FSGM \cite{Goodfellow2015ExplainingAH} and PGD \cite{madry2017towards}. FSGM and PGD attacks exploit the gradients of the neural network to build an adversarial image with goal of fooling the trained network. Our experiments on the lesion classification task show $+4.5\%$ and $+6.3\%$ higher robustness of RoS-KD than Baseline III on FSGM and PGD attacks, respectively.  Similarly, on the cardiopulmonary disease classification task, RoS-KD has  $+4.3\%$ and $+4.2\%$ higher robustness than Baseline III under FSGM and PGD attacks, respectively.

\subsection{Smoothness}
Introducing smoothness into the training paradigm of neural networks has been widely accepted as it is a technique to improve generalization and optimization. Smoothness can be implemented by replacing the activation functions, adding skip-connections in NNs \cite{He2016DeepRL,jaiswal2022training}, using soft labels replacing the hard labels \cite{szegedy2016rethinking}. In this work, we propose to enforce weight smoothness, by averaging multiple checkpoints along the training trajectory during the knowledge distillation. Our experiments in Table \ref{table:classification} illustrate the significant gain by the RoS-KD when we incorporate parameter averaging. To validate the induced smoothness, we plotted the counter plots of final loss landscape by Baseline III and RoS-KD using \cite{li2017visualizing}. Figure \ref{fig:loss_landscape} shows the comparison of counterplots of loss landscape of models trained with Baseline III and RoS-KD. We observed that RoS-KD has comparatively larger counter shape in the landscape with bigger basin for both lesion classification and cardiopulmonary classification, strengthening our claim of improved smoothness and better generalization of RoS-KD.

\section{Conclusion}
In this work, we propose a novel robust stochastic knowledge distillation framework (RoS-KD) which distills knowledge from multiple teacher networks trained on overlapping subsets of noisy labelled data to enhance deterrence to noise and improve generalization on unseen data. We additionally propose to incorporate  smoothing in the knowledge distillation step of RoS-KD,  which helps in flattening the global minima during the optimization, and improving generalization. Our extensive results on two popular real-world medical datasets demonstrate the effectiveness of RoS-KD, its state-of-the-art performance, and its robustness to adversarial attacks.

\section*{Acknowledgment}

This work is supported by the National Library of Medicine under Award No. 4R00LM013001 and National NSF AI Center at UT Austin. 

\bibliographystyle{./IEEEtran}
\bibliography{./IEEEexample}

\begin{thebibliography}{10}
\providecommand{\url}[1]{#1}
\csname url@samestyle\endcsname
\providecommand{\newblock}{\relax}
\providecommand{\bibinfo}[2]{#2}
\providecommand{\BIBentrySTDinterwordspacing}{\spaceskip=0pt\relax}
\providecommand{\BIBentryALTinterwordstretchfactor}{4}
\providecommand{\BIBentryALTinterwordspacing}{\spaceskip=\fontdimen2\font plus
\BIBentryALTinterwordstretchfactor\fontdimen3\font minus
  \fontdimen4\font\relax}
\providecommand{\BIBforeignlanguage}[2]{{%
\expandafter\ifx\csname l@#1\endcsname\relax
\typeout{** WARNING: IEEEtran.bst: No hyphenation pattern has been}%
\typeout{** loaded for the language `#1'. Using the pattern for}%
\typeout{** the default language instead.}%
\else
\language=\csname l@#1\endcsname
\fi
#2}}
\providecommand{\BIBdecl}{\relax}
\BIBdecl

\bibitem{wang2018tienet}
X.~Wang, Y.~Peng, L.~Lu, Z.~Lu, and R.~M. Summers, ``Tienet: Text-image
  embedding network for common thorax disease classification and reporting in
  chest x-rays,'' in \emph{CVPR}, 2018, pp. 9049--9058.

\bibitem{Wang_2017}
\BIBentryALTinterwordspacing
X.~Wang, Y.~Peng, L.~Lu, Z.~Lu, M.~Bagheri, and R.~M. Summers, ``Chestx-ray8:
  Hospital-scale chest x-ray database and benchmarks on weakly-supervised
  classification and localization of common thorax diseases,'' \emph{CVPR}, Jul
  2017. [Online]. Available: \url{http://dx.doi.org/10.1109/CVPR.2017.369}
\BIBentrySTDinterwordspacing

\bibitem{jaiswal2021scalp}
A.~Jaiswal, T.~Li, C.~Zander, Y.~Han, J.~F. Rousseau, Y.~Peng, and Y.~Ding,
  ``Scalp-supervised contrastive learning for cardiopulmonary disease
  classification and localization in chest x-rays using patient metadata,'' in
  \emph{2021 IEEE International Conference on Data Mining (ICDM)}.\hskip 1em
  plus 0.5em minus 0.4em\relax IEEE, 2021, pp. 1132--1137.

\bibitem{yao2017learning}
L.~Yao, E.~Poblenz, D.~Dagunts, B.~Covington, D.~Bernard, and K.~Lyman,
  ``Learning to diagnose from scratch by exploiting dependencies among
  labels,'' \emph{arXiv preprint arXiv:1710.10501}, 2017.

\bibitem{Liu_2019_ICCV}
J.~Liu, G.~Zhao, Y.~Fei, M.~Zhang, Y.~Wang, and Y.~Yu, ``Align, attend and
  locate: Chest x-ray diagnosis via contrast induced attention network with
  limited supervision,'' in \emph{ICCV}, Oct 2019.

\bibitem{seyyed2020chexclusion}
L.~Seyyed-Kalantari, G.~Liu, M.~McDermott, and M.~Ghassemi, ``Chexclusion:
  Fairness gaps in deep chest x-ray classifiers,'' \emph{arXiv preprint
  arXiv:2003.00827}, 2020.

\bibitem{han2021using}
Y.~Han, C.~Chen, L.~Tang, M.~Lin, A.~Jaiswal, S.~Wang, A.~Tewfik, G.~Shih,
  Y.~Ding, and Y.~Peng, ``Using radiomics as prior knowledge for thorax disease
  classification and localization in chest x-rays,'' in \emph{AMIA Annual
  Symposium Proceedings}, vol. 2021, 2021, p. 546.

\bibitem{algan2020label}
G.~Algan and I.~Ulusoy, ``Label noise types and their effects on deep
  learning,'' \emph{arXiv preprint arXiv:2003.10471}, 2020.

\bibitem{jaiswal2021radbert}
A.~Jaiswal, L.~Tang, M.~Ghosh, J.~F. Rousseau, Y.~Peng, and Y.~Ding,
  ``Radbert-cl: Factually-aware contrastive learning for radiology report
  classification,'' in \emph{Machine Learning for Health}.\hskip 1em plus 0.5em
  minus 0.4em\relax PMLR, 2021.

\bibitem{Fukuda2017EfficientKD}
T.~Fukuda, M.~Suzuki, G.~Kurata, S.~Thomas, J.~Cui, and B.~Ramabhadran,
  ``Efficient knowledge distillation from an ensemble of teachers,'' in
  \emph{INTERSPEECH}, 2017.

\bibitem{Wu2019MultiteacherKD}
M.-C. Wu, C.-T. Chiu, and K.-H. Wu, ``Multi-teacher knowledge distillation for
  compressed video action recognition on deep neural networks,'' \emph{ICASSP},
  pp. 2202--2206, 2019.

\bibitem{Yang2020ModelCW}
Z.~Yang, L.~Shou, M.~Gong, W.~Lin, and D.~Jiang, ``Model compression with
  two-stage multi-teacher knowledge distillation for web question answering
  system,'' \emph{WSDM}, 2020.

\bibitem{li2017visualizing}
H.~Li, Z.~Xu, G.~Taylor, C.~Studer, and T.~Goldstein, ``Visualizing the loss
  landscape of neural nets,'' \emph{arXiv preprint arXiv:1712.09913}, 2017.

\bibitem{hinton2015distilling}
G.~Hinton, O.~Vinyals, and J.~Dean, ``Distilling the knowledge in a neural
  network,'' \emph{arXiv preprint arXiv:1503.02531}, 2015.

\bibitem{yuan2021reinforced}
F.~Yuan, L.~Shou, J.~Pei, W.~Lin, M.~Gong, Y.~Fu, and D.~Jiang, ``Reinforced
  multi-teacher selection for knowledge distillation,'' in \emph{AAAI},
  vol.~35, no.~16, 2021, pp. 14\,284--14\,291.

\bibitem{johnson2019mimic}
A.~E. Johnson, T.~J. Pollard, N.~R. Greenbaum, and others., ``Mimic-cxr-jpg, a
  large publicly available database of labeled chest radiographs,'' \emph{arXiv
  preprint arXiv:1901.07042}, 2019.

\bibitem{garipov2018loss}
T.~Garipov, P.~Izmailov, D.~Podoprikhin, D.~Vetrov, and A.~G. Wilson, ``Loss
  surfaces, mode connectivity, and fast ensembling of dnns,'' in
  \emph{NeurIPS}, 2018, pp. 8803--8812.

\bibitem{izmailov2018averaging}
P.~Izmailov, D.~Podoprikhin, T.~Garipov, D.~Vetrov, and A.~G. Wilson,
  ``Averaging weights leads to wider optima and better generalization,''
  \emph{arXiv preprint arXiv:1803.05407}, 2018.

\bibitem{zhou2019multi}
S.~Zhou, Y.~Zhuang, and R.~Meng, ``Multi-category skin lesion diagnosis using
  dermoscopy images and deep cnn ensembles,'' \emph{l{\'\i}nea], ISIC
  Chellange}, 2019.

\bibitem{li2018thoracic}
Z.~Li, C.~Wang, M.~Han, Y.~Xue, W.~Wei, L.-J. Li, and L.~Fei-Fei, ``Thoracic
  disease identification and localization with limited supervision,'' in
  \emph{CVPR}, 2018, pp. 8290--8299.

\bibitem{wang2017chestx}
X.~Wang, Y.~Peng, L.~Lu, Z.~Lu, M.~Bagheri, and R.~M. Summers, ``Chestx-ray8:
  Hospital-scale chest x-ray database and benchmarks on weakly-supervised
  classification and localization of common thorax diseases,'' in \emph{CVPR},
  2017, pp. 2097--2106.

\bibitem{rajpurkar2017chexnet}
P.~Rajpurkar, J.~Irvin, K.~Zhu, B.~Yang, and others., ``Chexnet:
  Radiologist-level pneumonia detection on chest x-rays with deep learning,''
  2017.

\bibitem{kumar2018boosted}
P.~Kumar, M.~Grewal, and M.~M. Srivastava, ``Boosted cascaded convnets for
  multilabel classification of thoracic diseases in chest radiographs,'' in
  \emph{International Conference Image Analysis and Recognition}.\hskip 1em
  plus 0.5em minus 0.4em\relax Springer, 2018, pp. 546--552.

\bibitem{kurakin2016adversarial}
A.~Kurakin, I.~Goodfellow, and S.~Bengio, ``Adversarial machine learning at
  scale,'' \emph{arXiv preprint arXiv:1611.01236}, 2016.

\bibitem{Goodfellow2015ExplainingAH}
I.~J. Goodfellow, J.~Shlens, and C.~Szegedy, ``Explaining and harnessing
  adversarial examples,'' \emph{CoRR}, vol. abs/1412.6572, 2015.

\bibitem{madry2017towards}
A.~Madry, A.~Makelov, L.~Schmidt, and others., ``Towards deep learning models
  resistant to adversarial attacks,'' \emph{arXiv preprint arXiv:1706.06083},
  2017.

\bibitem{He2016DeepRL}
K.~He, X.~Zhang, S.~Ren, and J.~Sun, ``Deep residual learning for image
  recognition,'' \emph{2016 IEEE Conference on Computer Vision and Pattern
  Recognition (CVPR)}, pp. 770--778, 2016.

\bibitem{jaiswal2022training}
A.~K. Jaiswal, H.~Ma, T.~Chen, Y.~Ding, and Z.~Wang, ``Training your sparse
  neural network better with any mask,'' in \emph{International Conference on
  Machine Learning}.\hskip 1em plus 0.5em minus 0.4em\relax PMLR, 2022, pp.
  9833--9844.

\bibitem{szegedy2016rethinking}
C.~Szegedy, V.~Vanhoucke, S.~Ioffe, J.~Shlens, and Z.~Wojna, ``Rethinking the
  inception architecture for computer vision,'' in \emph{CVPR}, 2016, pp.
  2818--2826.

\end{thebibliography}

\end{document}